\documentclass[conference]{IEEEtran}
\usepackage{cite}
\usepackage{multirow}
\usepackage{makecell}
\usepackage[hyphens]{url}
\usepackage[hyphenbreaks]{breakurl}
\usepackage{subcaption}
\usepackage{amsmath,amssymb,amsfonts}
\usepackage{algorithmic}
\usepackage{graphicx}
\usepackage{textcomp}
\usepackage{xcolor}
\usepackage{array}
\usepackage{booktabs}
\usepackage{multirow}
\def\BibTeX{{\rm B\kern-.05em{\sc i\kern-.025em b}\kern-.08em
    T\kern-.1667em\lower.7ex\hbox{E}\kern-.125emX}}

\title{A Diversity-optimized Deep Ensemble Approach for Accurate Plant Leaf Disease Detection}

\author{
    \IEEEauthorblockN{Sai Nath Chowdary Medikonduru, Hongpeng Jin, Yanzhao Wu}
    \IEEEauthorblockA{
    \textit{Florida International University, Miami, FL, USA} \\
    \{smedi098, hjin008, yawu\}@fiu.edu}
}

\begin{document}
\maketitle
\begin{abstract}
Plant diseases pose a significant threat to global agriculture, causing over \$220 billion in annual economic losses and jeopardizing food security. The timely and accurate detection of these diseases from plant leaf images is critical to mitigating their adverse effects. Deep neural network Ensembles (Deep Ensembles) have emerged as a powerful approach to enhancing prediction accuracy by leveraging the strengths of diverse Deep Neural Networks (DNNs). However, selecting high-performing ensemble member models is challenging due to the inherent difficulty in measuring ensemble diversity. In this paper, we introduce the Synergistic Diversity (SQ) framework to enhance plant disease detection accuracy. First, we conduct a comprehensive analysis of the limitations of existing ensemble diversity metrics (denoted as Q metrics), which often fail to identify optimal ensemble teams. Second, we present the SQ metric, a novel measure that captures the synergy between ensemble members and consistently aligns with ensemble accuracy. Third, we validate our SQ approach through extensive experiments on a plant leaf image dataset, which demonstrates that our SQ metric substantially improves ensemble selection and enhances detection accuracy. Our findings pave the way for a more reliable and efficient image-based plant disease detection.
\end{abstract}
\section{Introduction}
\label{sec:intro}
Plant diseases pose a significant risk to global food security and agricultural productivity, causing annual economic losses exceeding \$220 billion~\cite{Bajait2020,kc-plant-pathology,review-detecting-plant-diseases}. The early detection and accurate classification of plant diseases from plant leaf images are crucial for mitigating their detrimental effects. Traditional diagnostic methods, which rely heavily on manual inspection by plant pathologists, are often labor-intensive, time-consuming, and prone to human errors~\cite{review-detecting-plant-diseases,golhani2018review}. With the growing demand for agricultural efficiency, the development of automated, scalable, and accurate plant disease detection solutions has become a critical need~\cite{fang2015current, debellis2021advances, mohanty2016using, too2019comparative}. 

Recent advances in machine learning, particularly in deep learning and computer vision, have revolutionized image-based analysis across various domains, including medical diagnostics~\cite{dl-medical-image-analysis,das2023privacy}, environmental science~\cite{WANG2022104110,shi2025deep}, and agricultural technologies~\cite{tian2020computer,dhanya2022deep}. Within the agricultural domain, plant disease detection from images poses unique challenges due to varying environmental conditions, differences in crop species, and the often subtle visual distinctions between healthy and diseased leaves. While deep neural networks (DNNs) have demonstrated considerable success in image recognition~\cite{he2016deep,huang2017densely,VT2021}, achieving high accuracy in real-world plant disease detection remains a critical challenge. This issue is exacerbated by factors such as variations in data quality, data imbalances across different diseases, and the need for robust generalization to unseen samples. 

Deep Neural Network Ensembles (\textit{Deep Ensembles}) have emerged as a promising approach for enhancing DNN prediction performance. By aggregating predictions from multiple diverse member DNNs, deep ensembles improve both accuracy and robustness, significantly outperforming a single DNN~\cite{deep-ensembles, ensemble-bigdata-2019, ensemble-icnc-2020, ganaie2021ensemble, Wu2023ModelLearning}. However, constructing efficient deep ensembles presents unique challenges, particularly the need to measure and optimize the diversity among ensemble members. High diversity is crucial, as it indicates that individual members can make uncorrelated errors with a high chance of complementing each other for improving overall ensemble performance~\cite{banfield2005ensemble,liu2019deep,EnsembleBenchCVPR,sq-diversity-cogmi}. Despite its importance, measuring and leveraging diversity in deep ensembles remains an open research challenge, especially for large-scale real-world datasets.

In this paper, we introduce the \textit{Synergistic Diversity (SQ)}, a novel approach to optimizing ensemble diversity for accurate plant disease detection. We make three original contributions. 
\textit{First}, we provide a comprehensive analysis of the existing ensemble diversity metrics, referred to as Q metrics, and show their inherent limitations of low correlation with ensemble accuracy. 
\textit{Second}, we present a novel ensemble diversity metric, the \textit{SQ metric}, which captures the synergy among ensemble member models. 
Unlike the traditional Q metrics, which often fail to recognize high-quality ensembles with complementary member models, our SQ metric closely correlates with ensemble accuracy, providing an effective method for selecting high-quality ensembles with high accuracy. 
\textit{Third}, extensive experiments on plant leaf disease datasets demonstrate that our SQ approach not only outperforms existing Q diversity measures in enhancing ensemble accuracy but also establishes a scalable and robust mechanism to construct high-quality ensembles. This paper demonstrates the high potential of our SQ framework to enhance image-based plant disease detection and contribute to the broader adoption of deep ensemble techniques in agricultural applications.

\section{Related Work}
\label{sec:related}

Deep learning-powered image classification has enabled the detection of plant diseases from plant leaf images~\cite{ferentinos2018deep,saleem2019plant,Arsenovic2019,li2021plant,dhaka2023role}. However, most existing studies focus on using a single deep neural network, which may not perform well across different plant species or under varying environmental conditions~\cite{Bajait2020,Arsenovic2019,shoaib2023advanced}. Several early studies have explored ensemble learning to enhance plant leaf disease detection~\cite{Bajait2020, ganaie2021ensemble, plant-disease-detection-ensemble-learning, plant-leaf-disease-detection-ens-xai, CATALREIS2024109790}. However, these studies often leverage all available models to build ensemble teams, which can result in high computational costs and potentially suboptimal accuracy~\cite{ganaie2021ensemble, EnsembleBenchCogMI}. To address these issues, there is a critical need for efficient ensemble strategies. Diversity-based ensemble selection is a popular technique for reducing ensemble execution costs by selecting only a subset of available models to construct the ensemble~\cite{pang2019improving,wu2024hierarchical}. Existing diversity metrics, such as Cohen’s Kappa (CK)~\cite{Cohen1960}, Q Statistics (QS)~\cite{Yule1900}, Binary Disagreement (BD)~\cite{skalak1996sources}, and Generalized Diversity (GD)~\cite{partridge1997software}, referred to as Q metrics, are commonly used to guide the ensemble selection process~\cite{kuncheva2003measures, liu2019improving}. While these Q metrics provide valuable insights into the diversity of ensemble member models, they fail to precisely identify high-quality ensembles due to their low correlations to ensemble performance~\cite{kuncheva2003measures,EnsembleBenchCVPR,sq-diversity-cogmi}. This limitation also impedes their effective application in plant disease detection~\cite{golhani2018review, EnsembleBenchCVPR, plant-disease-detection-ensemble-learning,plant-leaf-disease-detection-ens-xai}. We present our Synergistic Diversity (SQ) metric to address this problem, providing highly accurate and efficient deep ensembles for enhancing plant disease detection performance.

\section{Methodology}
\label{sec:methodology}

This paper aims to enhance the prediction performance of deep neural network ensembles for plant disease detection by optimizing ensemble diversity. Specifically, it presents a systematic comparison between traditional Q-diversity metrics and our Synergistic Diversity (SQ) metric. The SQ metric addresses the limitations of existing Q metrics by effectively capturing complementary prediction capabilities among ensemble members, substantially improving ensemble accuracy.

\begin{figure}[h!]
    \centering
\includegraphics[width=0.43\textwidth]
    {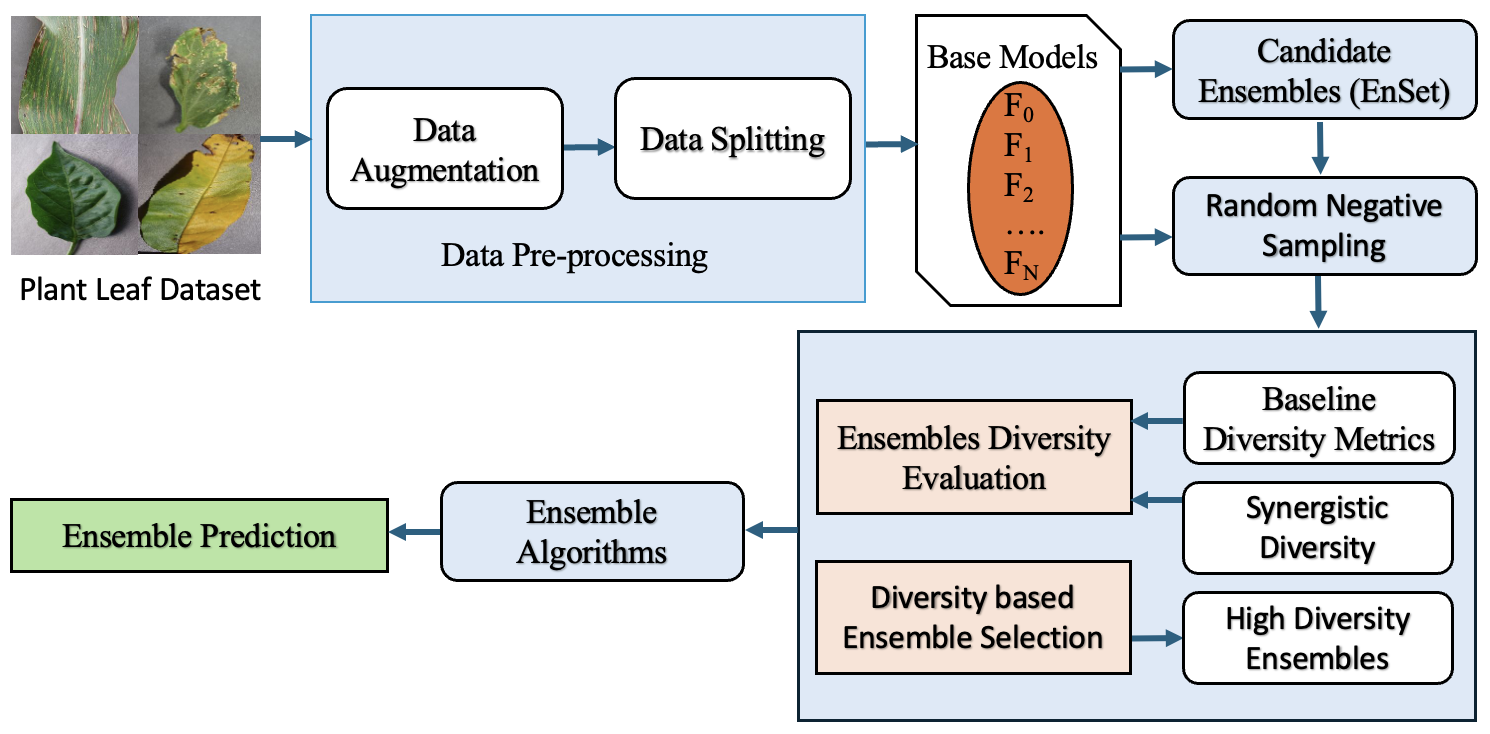}
    \caption{Overview of Our Deep Ensemble Approach}
    \label{fig:ProcessDiagram}
\end{figure}

Figure~\ref{fig:ProcessDiagram} illustrates the overview of our deep ensemble framework. Our approach follows six main steps. 
\textit{First}, we perform data pre-processing on the Plant Leaf Dataset~\cite{dobrovsky2023dataset,malviya2023dataset}, which consists of labeled images of healthy and diseased plant leaves. 
\textit{Second,} we train multiple base models to form the base model pool. 
\textit{Third,} we generate possible ensembles from the base model pool and include them in the set of candidate ensembles ($EnsSet$).
\textit{Fourth,} we evaluate the ensemble diversity for these candidate ensembles.
\textit{Fifth}, we select high-quality ensembles with high ensemble diversity. 
\textit{Sixth}, we use ensemble consensus methods to aggregate the individual member model predictions to produce the ensemble predictions. 
We describe these steps in detail below. 

\noindent \textbf{(1) Data Pre-processing:} To enhance the robustness and generalizability of the trained models, we pre-process the dataset with (i) \textit{data augmentation} techniques, such as random rotation, flipping, cropping, and color adjustments, which mitigate overfitting by simulating a variety of real-world conditions, and (ii) \textit{data splitting}, where the preprocessed dataset is split into training, validation, and testing subsets in proportions of 80\%, 10\%, and 10\%, respectively.

\noindent \textbf{(2) Base Model Training:} A pool of deep neural networks, referred to as base models (\(F_0, F_1, \dots, F_N\)), is independently trained on the preprocessed dataset. These models utilize state-of-the-art deep neural network architectures, such as ResNet~\cite{he2016deep} and DenseNet~\cite{huang2017densely}, to extract discriminative features for classifying plant diseases. These trained base models form the base model pool for ensemble construction.

\noindent \textbf{(3) Candidate Ensemble Construction:} 
Using these trained base models, a set of candidate ensembles ($EnsSet$) can be generated. These ensembles form the basis for diversity evaluation and performance assessment.

\noindent \textbf{(4) Ensemble Diversity Evaluation:} We first follow~\cite{EnsembleBenchCogMI} to sample negative samples where one or multiple base models make prediction errors. We then evaluate ensemble diversity using various diversity metrics, such as existing Q diversity metrics (e.g., CK, BD, GD, and KW) and our proposed SQ diversity metric on these negative samples. To ensure consistency, we calculate CK diversity scores using 1-CK so that high diversity scores correspond to greater ensemble diversity. 
The formal definitions of these Q-diversity metrics are comprehensively detailed in~\cite{EnsembleBenchCogMI}.

\noindent \textbf{(5) Ensemble Selection:} We rank candidate ensembles based on their diversity scores. In the case of ties, smaller ensembles are given higher priority in the ranking, given their lower ensemble costs. The top-K ensembles are then selected for further evaluation to identify the optimal ensembles.

\noindent \textbf{(6) Ensemble Consensus:} The ensemble predictions are generated using an ensemble consensus method. This process aggregates the prediction results of each individual member model within each selected ensemble. Popular methods include majority voting, which chooses the class with the most votes, and soft voting, which averages the predicted probabilities across ensemble members to produce the ensemble predictions. As suggested by~\cite{EnsembleBenchCogMI}, we use soft voting by default. 

\begin{figure*}[h!]
    \centering
    \begin{subfigure}[t]{0.245\textwidth}
        \includegraphics[width=\textwidth]{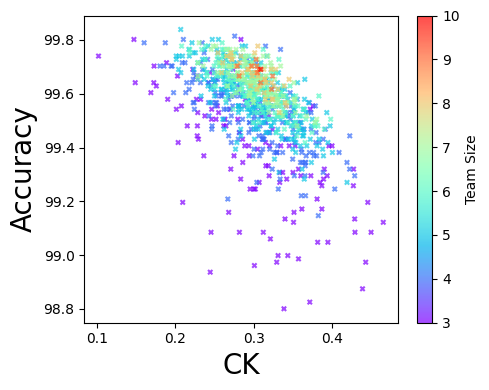}
        \caption{Cohen's Kappa (CK)}
        \label{fig:Q_Diversity_CK}
    \end{subfigure}
    \hfill
    \begin{subfigure}[t]{0.245\textwidth}
        \includegraphics[width=\textwidth]{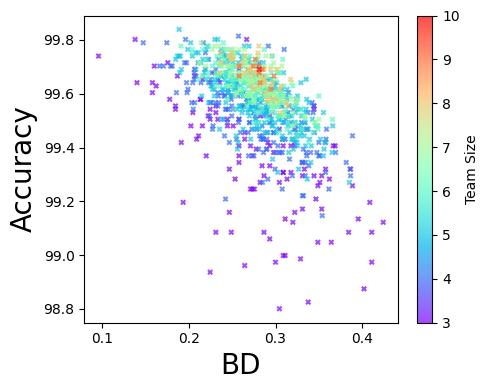}
        \caption{Binary Disagreement (BD)}
        \label{fig:Q_Diversity_BD}
    \end{subfigure}
    \hfill
        \begin{subfigure}[t]{0.245\textwidth}
        \includegraphics[width=\textwidth]{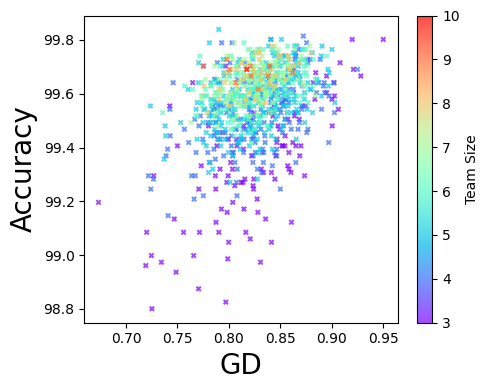}
        \caption{Generalized Diversity (GD)}
        \label{fig:Q_Diversity_GD}
    \end{subfigure}
    \hfill
    \begin{subfigure}[t]{0.245\textwidth}
        \includegraphics[width=\textwidth]{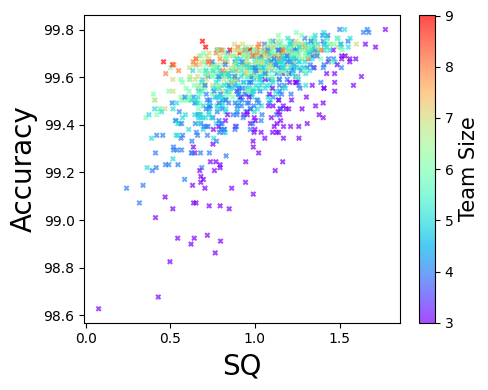}
        \caption{Synergistic Diversity (SQ)}
        \label{fig:Q_Diversity_SQ}
    \end{subfigure}
    \caption{Scatter Plots of Q Metric Scores and Ensemble Accuracy on The Plant Leaf Dataset: \textit{the legend on the right shows how different colors correspond to different team sizes}.}
    \label{fig:QmetricsvsSQ}
\end{figure*}

\subsection{Limitations of Q-Diversity Metrics}

Figure~\ref{fig:QmetricsvsSQ} shows the ensemble diversity and accuracy for candidate ensembles with different team sizes marked in different colors. We observe that traditional Q diversity metrics, such as CK, BD, and GD, fail to show a clear correlation with ensemble accuracy. Higher Q diversity scores may not correspond to higher ensemble accuracy, showing that using these Q diversity metrics may not effectively select high-quality ensembles.  
These limitations motivate us to develop the SQ metric to optimize ensemble diversity measurements and achieve stronger correlations with ensemble accuracy.

\subsection{Synergistic Diversity (SQ) Metric}

The Synergistic Diversity (SQ) metric is designed to address the limitations of traditional Q metrics by providing a more comprehensive evaluation of ensemble diversity. Unlike Q metrics, which rely solely on the disagreement of ensemble members on negative samples, SQ captures the complementary prediction capabilities among ensemble members, using both the disagreement between ensemble member models and a focal model and the agreements among non-focal models on the negative samples from the focal model. 
We adopt the focal model concept from~\cite{EnsembleBenchCVPR}, where one ensemble member model is designated as the focal model. Negative samples are then randomly sampled as those on which the focal model makes prediction errors. The remaining ensemble member models are referred to as non-focal models. 
The SQ metric measures ensemble diversity through two key components.

\noindent \textbf{Disagreement between Ensemble Members and The Focal Model (\( \text{SQ-}\epsilon \)):} This component measures the extent to which non-focal models compensate for the errors made by a focal model. The concept of Binary Disagreement (BD)~\cite{skalak1996sources} is adopted to measure the disagreement between the ensemble members and the focal model on the negative samples where the focal model made mistakes, i.e., its hard cases. By promoting the disagreement on these challenging samples, a higher \( \text{SQ-}\epsilon \) value indicates a higher chance for non-focal models to correct the focal model mistakes to improve overall accuracy. 

\noindent \textbf{Agreement among Non-focal Models (\( \text{SQ-}\alpha \)):} 
This component captures the synergy within the ensemble by measuring the agreement among non-focal models on the negative samples where the focal model makes mistakes. Agreement is quantified using Cohen’s Kappa (CK)~\cite{Cohen1960} to assess how consistently these non-focal models align in their predictions. A higher \( \text{SQ-}\alpha \) value reflects that these non-focal models possess more substantial capabilities to complement the focal model, increasing the chance of correct ensemble predictions.

The final SQ score is computed by integrating these two components:
\begin{equation} 
    \text{SQ} = w_\epsilon \cdot \text{SQ-}\epsilon + w_\alpha \cdot \text{SQ-}\alpha,
\end{equation}
where \( w_\epsilon \) and \( w_\alpha \) are the weights that control the relative importance of disagreement and agreement, both of which are set to 1 by default. 
We validate the effectiveness of our SQ metric on the Plant Leaf Dataset, as shown in Figure~\ref{fig:Q_Diversity_SQ}. It shows a strong positive correlation between SQ scores and ensemble accuracy, demonstrating that our SQ metric can be effectively used in identifying high-quality ensembles.

\begin{table}[h!]
\caption{Performance of Individual Base Models}
\label{table:accuracies}
\small
\centering
\begin{tabular}{|c|c|c|}
\hline
\textbf{Model ID} & \textbf{Model Name} & \textbf{Accuracy (\%)} \\ \hline
0 & DenseNet169       & 96.41 \\ \hline
1 & DenseNet121       & \textbf{99.15} \\ \hline
2 & EfficientNetB3    & 98.52 \\ \hline
3 & EfficientNetB4    & \underline{98.73} \\ \hline
4 & DenseNet201       & 97.80 \\ \hline
5 & ResNet50          & 97.49 \\ \hline
6 & ResNet101         & 96.70 \\ \hline
7 & ResNet152         & 96.94 \\ \hline
8 & ViT B-16          & 96.88 \\ \hline
9 & FineTuned ViT     & 98.48 \\
\hline
\end{tabular}
\end{table}
\begin{table*}[h!]
\caption{Top 10 Ensembles for Each Diversity Metric: \textit{our SQ metric effectively identifies high-quality ensemble teams, ranking first for 7 out of 10 and second for the remaining 3, significantly outperforming other diversity metrics}.}
\label{table:topensembles}
\small
\centering
\setlength{\tabcolsep}{3pt}
\renewcommand{\arraystretch}{1.1}
\resizebox{\textwidth}{!}{
\begin{tabular}{|c|c|c|c|c|c|c|c|c|c|c|c|}
\hline
\textbf{Diversity Metrics} & \textbf{Evaluation Metrics} & \textbf{Top1} & \textbf{Top2} & \textbf{Top3} & \textbf{Top4} & \textbf{Top5} & \textbf{Top6} & \textbf{Top7} & \textbf{Top8} & \textbf{Top9} & \textbf{Top10} \\
\hline
\multirow{4}{*}{CK} 
& Ensemble Team & 068 & 0678 & 0467 & 067 & 0568 & 02678 & 078 & 678 & 0468 & 467 \\
& Ensemble Acc (\%) & 99.05 & 99.28 & 99.22 & 98.82 & 99.29 & 99.49 & 99.17 & 99.07 & 99.41 & 98.92 \\
& Best Single Acc (\%) & 96.88 & 96.94 & 97.80 & 96.94 & 97.49 & 98.52 & 96.94 & 96.94 & 97.80 & 97.80 \\
& Acc Improvement (\%) & 2.17 & 2.34 & 1.42 & 1.88 & 1.80 & 0.97 & 2.23 & 2.13 & 1.61 & 1.12 \\
\hline
\multirow{4}{*}{BD} 
& Ensemble Team & 068 & 0678 & 067 & 02678 & 0467 & 0568 & 568 & 678 & 078 & 05678 \\
& Ensemble Acc (\%) & 99.05 & 99.28 & 98.82 & 99.49 & 99.22 & 99.29 & 99.18 & 99.07 & 99.17 & 99.36 \\
& Best Single Acc (\%) & 96.88 & 96.94 & 96.94 & 98.52 & 97.80 & 97.49 & 97.49 & 96.94 & 96.94 & 97.49 \\
& Acc Improvement (\%) & 2.17 & 2.34 & 1.88 & 0.97 & 1.42 & 1.80 & 1.69 & 2.13 & 2.23 & 1.87 \\
\hline
\multirow{4}{*}{GD} 
& Ensemble Team & 023 & 13578 & 016 & 13679 & 1236 & 3479 & 1358 & 2468 & 1238 & 0238 \\
& Ensemble Acc (\%) & \underline{99.58} & \underline{99.74} & 99.41 & \textbf{99.75} & \underline{99.72} & \underline{99.72} & \textbf{99.76} & \underline{99.47} & \textbf{99.78} &  \underline{99.67} \\
& Best Single Acc (\%) & 98.73 & 99.15 & 99.15 & 99.15 & 99.15 & 98.73 & 99.15 & 98.52 & 99.15 & 98.73 \\
& Acc Improvement (\%) & 0.85 & 0.59 & 0.26 & 0.60 & 0.57 & 0.99 & 0.61 & 0.95 & 0.63 & 0.94 \\
\hline
\multirow{4}{*}{KW} 
& Ensemble Team & 02678 & 0678 & 0245678 & 045678 & 056789 & 04678 & 05678 & 04568 & 0467 & 06789 \\
& Ensemble Acc (\%) & 99.49 & 99.28 & \underline{99.47} & 99.44 & 99.47 & 99.34 & 99.36 & 99.43 & 99.22 & 99.46 \\
& Best Single Acc (\%) & 98.52 & 96.94 & 98.52 & 97.80 & 98.48 & 97.80 & 97.49 & 97.80 & 97.80 & 98.48 \\
& Acc Improvement (\%) & 0.97 & 2.34 & 0.95 & 1.64 & 0.99 & 1.54 & 1.87 & 1.63 & 1.42 & 0.98 \\
\hline
\multirow{4}{*}{\makecell{SQ\\(our approach)}} 
& Ensemble Team & 139 & 1239 & 1379 & 138 & 12379 & 12378 & 01239 & 12359 & 1349 & 1359 \\
& Ensemble Acc (\%) & \textbf{99.80} & \textbf{99.79} & \textbf{99.80} & \underline{99.73} & \textbf{99.76} & \textbf{99.74} & \underline{99.74} & \textbf{99.80} & \underline{99.75} & \textbf{99.79} \\
& Best Single Acc (\%) & 99.15 & 99.15 & 99.15 & 99.15 & 99.15 & 99.15 & 99.15 & 99.15 & 99.15 & 99.15 \\
& Acc Improvement (\%) & 0.65 & 0.64 & 0.65 & 0.58 & 0.61 & 0.59 & 0.59 & 0.65 & 0.60 & 0.64 \\
\hline
\end{tabular}
}
\end{table*}

\section{Experimental Analysis}
\label{sec:experiments}

\noindent \textbf{Experimental Setup:} 
All experiments are conducted on a server with an Intel Xeon E5-1620 CPU and an NVIDIA Tesla T4 GPU. 
The dataset used in this study consisted of approximately 80,000 labeled plant leaf images, capturing a diverse range of healthy and diseased conditions across multiple crop species~\cite{dobrovsky2023dataset,malviya2023dataset}. We preprocessed the dataset using popular data augmentation techniques, including rotation, flipping, and cropping. The dataset was then split into training (80\%), validation (10\%), and testing (10\%) sets. State-of-the-art deep learning models, including DenseNet~\cite{huang2017densely}, EfficientNet~\cite{Efficientnet}, ResNet~\cite{he2016deep}, and Vision Transformers (ViTs)~\cite{VT2021} are trained independently on the preprocessed dataset to form the base model pool. These models are chosen for their high accuracy in image classification and diverse neural network architectures. 
Table~\ref{table:accuracies} presents the base model pool with each individual base model and its accuracy. 
DenseNet121 achieved the highest accuracy of 99.15\%, while other base models also achieved comparable accuracy, providing a solid foundation for constructing and evaluating ensemble teams.

\begin{table*}[h!]
\caption{Image Examples from the Plant Leaf Disease Dataset with Member Model and Ensemble Predictions}
\label{table:results}
\small
\centering
\resizebox{\textwidth}{!}{
\begin{tabular}{|c|c|c|c|}
\hline
\textbf{Image} & 
\includegraphics[width=3cm]{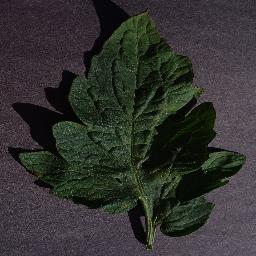} & 
\includegraphics[width=3cm]{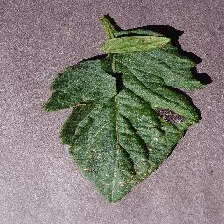} & 
\includegraphics[width=3cm]{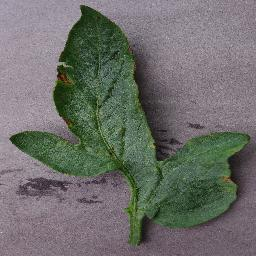} \\
\hline
\textbf{Ground Truth Label} & 
\textit{Tomato\_Target\_Spot} & 
\textit{Tomato\_Early\_blight} & 
\textit{Tomato\_Bacterial\_Spot} \\
\hline
\makecell{\textbf{$F_2F_5F_9$}\\{(Accuracy: 99.49\%)}} & 
\includegraphics[height=3cm, width=4.5cm]{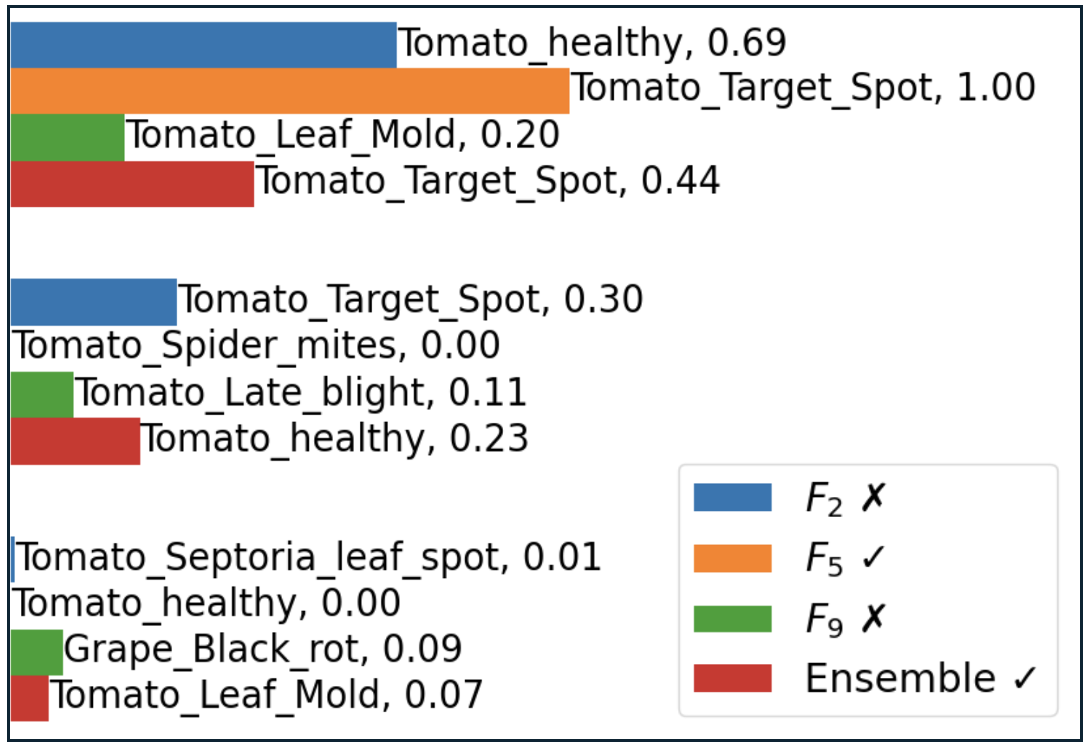} & 
\includegraphics[height=3cm, width=4.5cm]{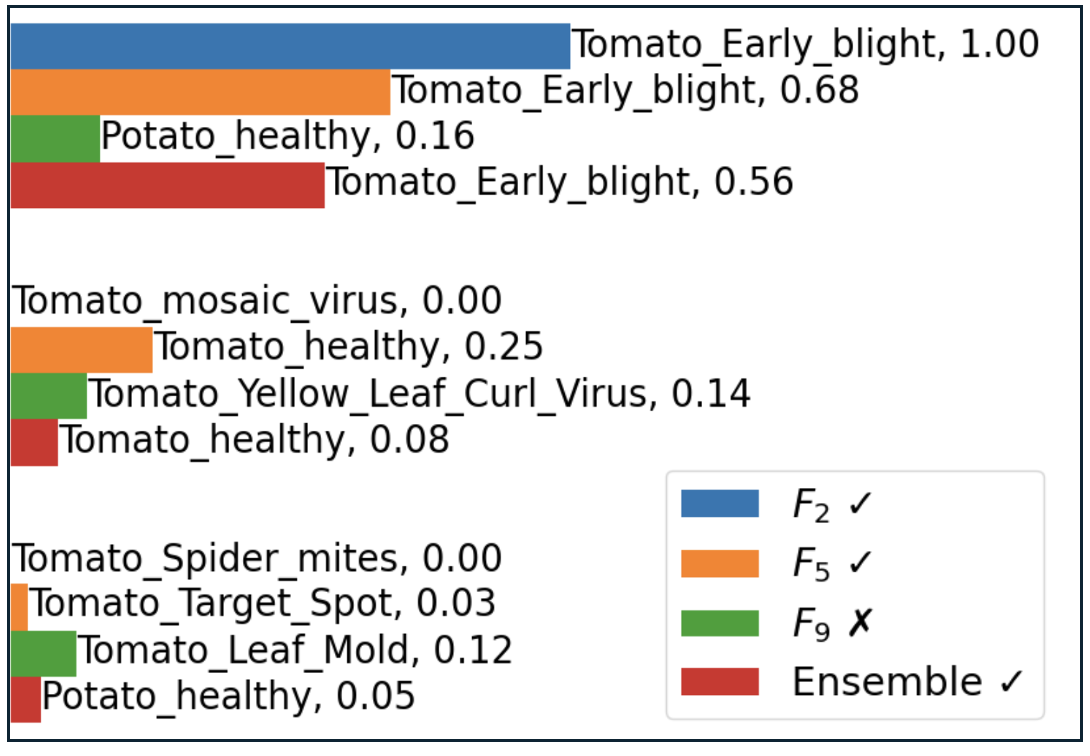} & 
\includegraphics[height=3cm, width=4.5cm]{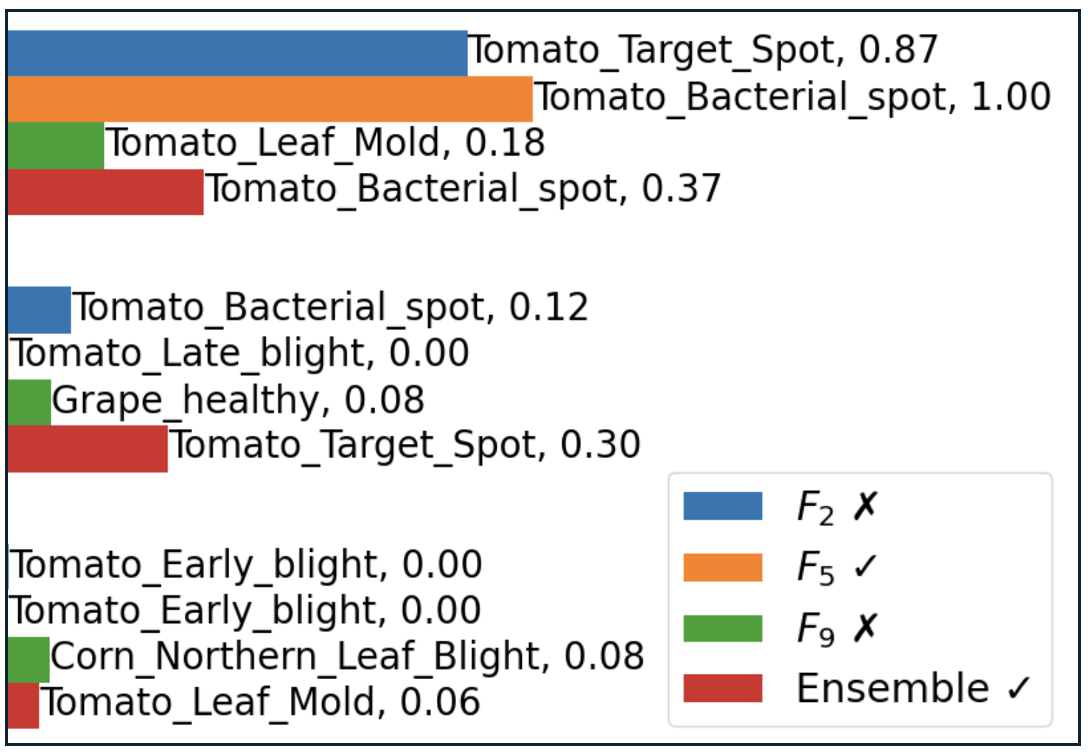} \\
\hline
\makecell{\textbf{$F_3F_6F_8$}\\{(Accuracy: 99.58\%)}} & 
\includegraphics[height=3cm, width=4.5cm]{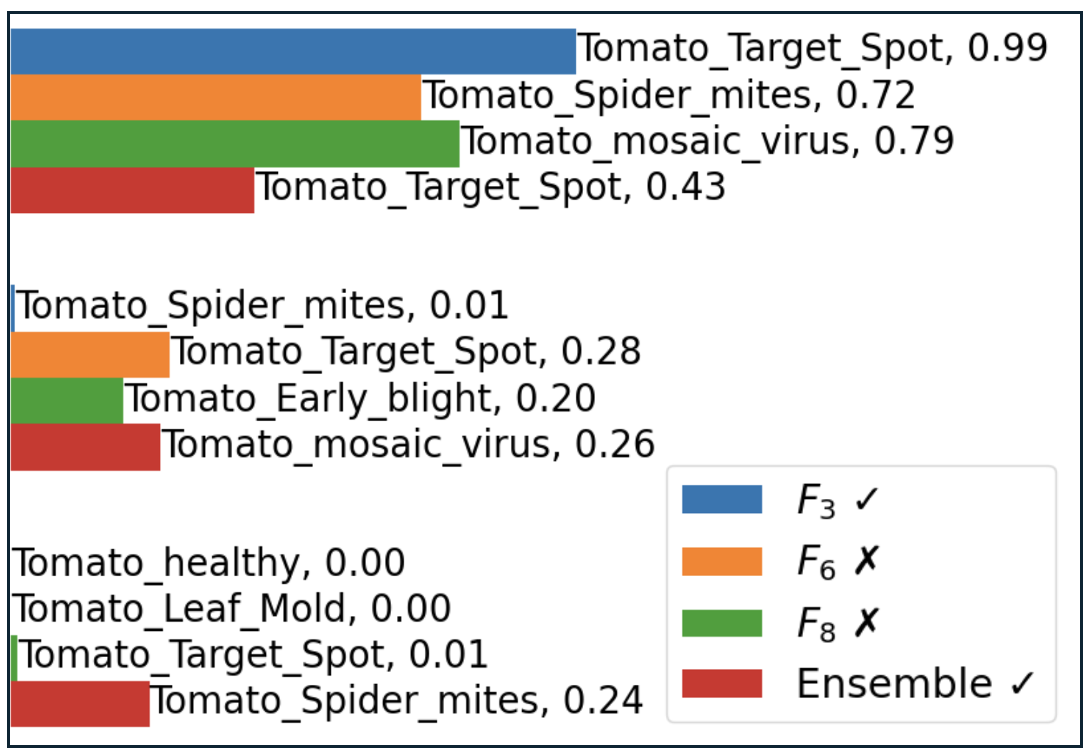} & 
\includegraphics[height=3cm, width=4.5cm]{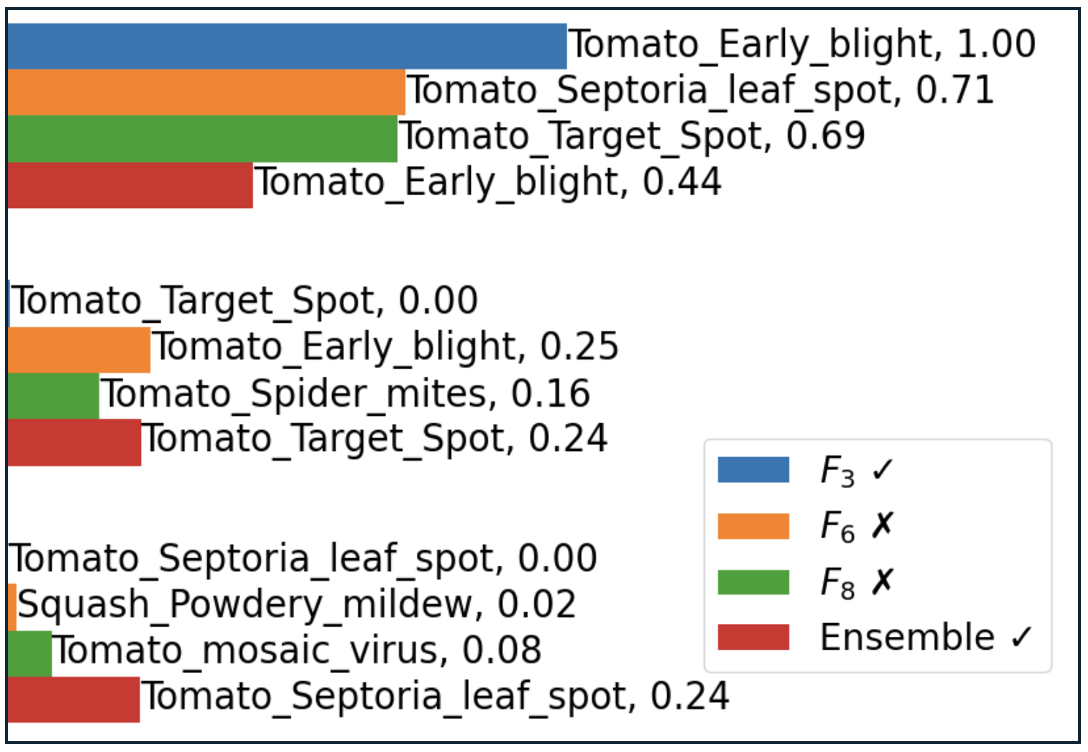} & 
\includegraphics[height=3cm, width=4.5cm]{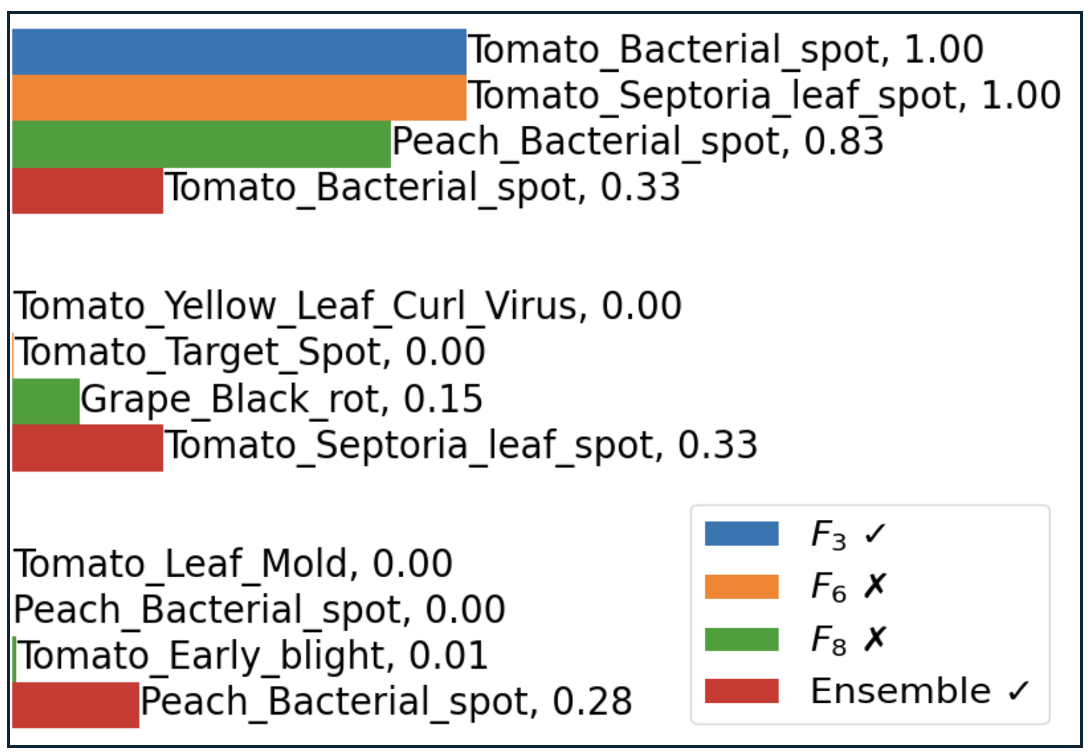} \\
\hline
\textbf{Predicted Truth Label} & 
\textit{Tomato\_Target\_Spot} & 
\textit{Tomato\_Early\_blight} & 
\textit{Tomato\_Bacterial\_Spot} \\
\hline
\end{tabular}}
\end{table*}

\subsection{Ensemble Performance Evaluation}
We evaluate these candidate ensembles in $EnsSet$ using traditional Q diversity metrics, including Cohen’s Kappa (CK), Binary Disagreement (BD), Generalized Diversity (GD), and Kohavi-Wolpert Variance (KW), as well as the proposed Synergistic Diversity (SQ) metric. Table~\ref{table:topensembles} presents the Top 10 ensembles selected by each diversity metric, including their accuracy and improvements over their best-performing individual member model. The highest accuracy for each rank is highlighted in bold, with the second highest underlined. 
We highlight three interesting observations. 
\textit{First}, our proposed SQ metric achieved the highest accuracy of 99.80\% with the ensemble team \texttt{139}, which consists of only three member models with relatively low ensemble execution costs. Notably, this ensemble team outperforms the best single model (DenseNet121 with 99.15\% accuracy) by 0.65\%. 
\textit{Second}, for the Top-3 ranked ensembles, our SQ metric effectively identified the best-performing ensembles for Top-1, Top-2, and Top-3 rankings, all achieving ensemble accuracy above 99.79\%.
This demonstrates its effectiveness in identifying high-quality ensembles with high ensemble accuracy.
\textit{Third,} in comparison, while Generalized Diversity (GD) can still identify the best performing ensembles for the Top-4, Top-7, and Top-9 rankings, other Q diversity metrics, including Cohen’s Kappa (CK), Binary Disagreement (BD), and Kohavi-Wolpert Variance (KW), fail to identify the ensembles with accuracy above 99.50\% among their Top-10 ensembles, showing their inherent limitations in effectively capturing the complementary prediction capabilities among ensemble member models and delivering consistent performance.

\subsection{Correlation Analysis}
\begin{figure}[h!]
    \centering
\includegraphics[width=0.45\textwidth]
    {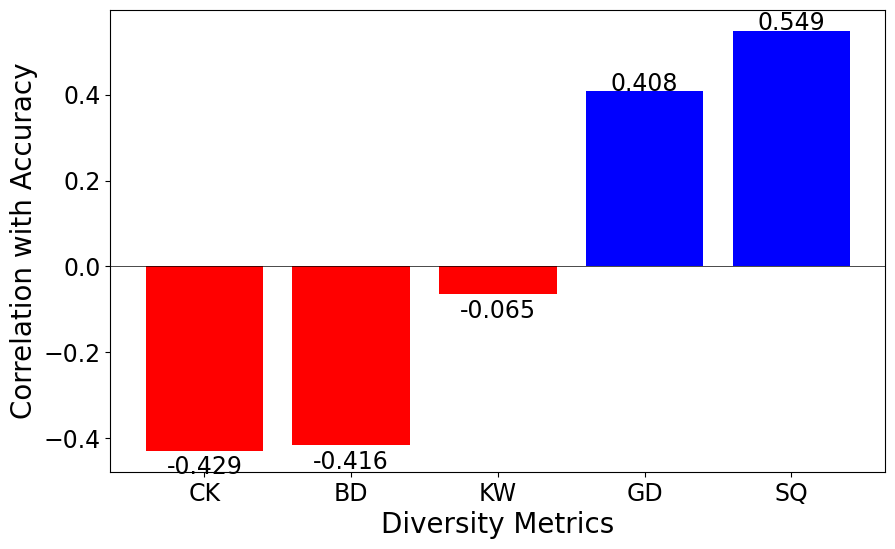}
    \caption{Correlation of Ensemble Diversity and Accuracy.}
    \label{fig:Correlation}
\end{figure}

We further investigate the correlations between these ensemble diversity metrics and ensemble accuracy as shown in Figure~\ref{fig:Correlation}. 
We observed that our proposed Synergistic Diversity (SQ) metric exhibited the strongest positive correlation with accuracy, i.e., 0.549, followed by GD of 0.408. The other traditional Q metrics, including CK, BD, and KW, exhibited negative correlations, i.e., -0.429, -0.416, and -0.065, respectively. This indicates that higher CK, BD or KW diversity scores may not necessarily translate to improved ensemble accuracy. 
This finding further explains the enhanced ensemble performance by our SQ approach.

\subsection{Case Studies}
We conduct several case studies to visualize the prediction results by ensemble member models to validate our SQ metric. 
Table~\ref{table:results} presents a comparison of predictions made by individual member models and SQ-based ensembles across three representative disease categories: \textit{Tomato Target Spot}, \textit{Tomato Early Blight}, and \textit{Tomato Bacterial Spot}. 
The results show that SQ-based ensembles can effectively correct prediction errors made by individual member models. 
For example, model $F_2$ (EfficientNetB3) and $F_9$ (FineTuned ViT) made different mistakes for the first example, misclassifying it as \texttt{Tomato\_healthy} and \texttt{Tomato\_mosaic\_virus}, respectively. However, model $F_9$ exhibited high confidence in predicting the correct class, \texttt{Tomato\_Target\_Spot}, enabling the ensemble  $F_2F_5F_9$ to make the correct prediction. Similar observations can be found for the other examples in Table~\ref{table:results}. 
These findings show that our SQ metric can be applied effectively in optimizing ensemble performance in real-world scenarios, particularly in high-stakes applications like plant disease detection, where accurate and reliable predictions are essential for mitigating agricultural losses.

\section{Conclusion} \label{sec:conclusion}
We make three original contributions to enhance image-based plant disease detection using deep ensembles. 
\textit{First}, we present a comprehensive study of existing Q diversity metrics and demonstrate their inherent limitations in capturing the complementary predictive capabilities among ensemble member models. 
\textit{Second}, we introduce a novel Synergistic Diversity (SQ) framework to effectively measure the synergistic effects among ensemble member models, which is closely correlated to ensemble accuracy.
\textit{Third}, we conduct comprehensive experiments on the plant leaf image dataset, which demonstrate that our SQ metric can effectively identify high-quality ensembles for detecting plant diseases and significantly outperform the existing Q metrics with high accuracy.

\section*{Acknowledgment}
The authors acknowledge the National Artificial Intelligence Research Resource (NAIRR) Pilot (NAIRR240244) and Amazon Web Services for partially contributing to this research. 
Any opinions, findings, and conclusions or recommendations expressed in this material are those of the author(s) and do not necessarily reflect the views of the funding agency and company mentioned above.

\bibliographystyle{IEEEtran}
\small{
\bibliography{reference}

@INPROCEEDINGS{plant-disease-detection-ensemble-learning,
  author={Kondaveeti, Hari Kishan and Ujini, Kalyan Gandhi and Pavankumar, Bikkina Veera Venkata and Tarun, Bollu Sai and Gopi, Simhadri Chinna},
  booktitle={2023 2nd International Conference on Computational Systems and Communication (ICCSC)}, 
  title={Plant Disease Detection Using Ensemble Learning}, 
  year={2023},
  volume={},
  number={},
  pages={1-6},
  doi={10.1109/ICCSC56913.2023.10142982}}

@ARTICLE{plant-leaf-disease-detection-ens-xai,
  author={Oad, Ammar and Abbas, Syed Shoaib and Zafar, Amna and Akram, Beenish Ayesha and Dong, Feng and Talpur, Mir Sajjad Hussain and Uddin, Mueen},
  journal={IEEE Access}, 
  title={Plant Leaf Disease Detection Using Ensemble Learning and Explainable AI}, 
  year={2024},
  volume={12},
  number={},
  pages={156038-156049},
  doi={10.1109/ACCESS.2024.3484574}}

@article{CATALREIS2024109790,
title = {Integrated deep learning and ensemble learning model for deep feature-based wheat disease detection},
journal = {Microchemical Journal},
volume = {197},
pages = {109790},
year = {2024},
issn = {0026-265X},
doi = {https://doi.org/10.1016/j.microc.2023.109790},
url = {https://www.sciencedirect.com/science/article/pii/S0026265X23014091},
author = {Hatice {Catal Reis} and Veysel Turk}
}

@article{tian2020computer,
title = {{Computer vision technology in agricultural automation — A review}},
journal = {Information Processing in Agriculture},
volume = {7},
number = {1},
pages = {1-19},
year = {2020},
issn = {2214-3173},
author = {Hongkun Tian and Tianhai Wang and Yadong Liu and Xi Qiao and Yanzhou Li}
}

@article{dhanya2022deep,
title = {Deep learning based computer vision approaches for smart agricultural applications},
journal = {Artificial Intelligence in Agriculture},
volume = {6},
pages = {211-229},
year = {2022},
issn = {2589-7217},
author = {V.G. Dhanya and A. Subeesh and N.L. Kushwaha and Dinesh Kumar Vishwakarma and T. {Nagesh Kumar} and G. Ritika and A.N. Singh}
}

@INPROCEEDINGS{das2023privacy,
  author={Das, Badhan Chandra and Hadi Amini, M. and Wu, Yanzhao},
  booktitle={2023 IEEE International Conference on Bioinformatics and Biomedicine (BIBM)}, 
  title={Privacy Risks Analysis and Mitigation in Federated Learning for Medical Images}, 
  year={2023},
  volume={},
  number={},
  pages={1870-1873},
  doi={10.1109/BIBM58861.2023.10385829}}

@INPROCEEDINGS{ensemble-bigdata-2019,
  author={Chow, Ka-Ho and Wei, Wenqi and Wu, Yanzhao and Liu, Ling},
  booktitle={2019 IEEE International Conference on Big Data (Big Data)}, 
  title={Denoising and Verification Cross-Layer Ensemble Against Black-box Adversarial Attacks}, 
  year={2019},
  volume={},
  number={},
  pages={1282-1291},
  doi={10.1109/BigData47090.2019.9006090}}

@article{shi2025deep,
  title={Deep learning and foundation models for weather prediction: A survey},
  author={Shi, Jimeng and Shirali, Azam and Jin, Bowen and Zhou, Sizhe and Hu, Wei and Rangaraj, Rahuul and Wang, Shaowen and Han, Jiawei and Wang, Zhaonan and Lall, Upmanu and others},
  journal={arXiv preprint arXiv:2501.06907},
  year={2025}
}

@article{WANG2022104110,
title = {A comprehensive review on deep learning based remote sensing image super-resolution methods},
journal = {Earth-Science Reviews},
volume = {232},
pages = {104110},
year = {2022},
issn = {0012-8252},
author = {Peijuan Wang and Bulent Bayram and Elif Sertel}
}

@INPROCEEDINGS{ensemble-icnc-2020,
  author={Wei, Wenqi and Liu, Ling and Loper, Margaret and Chow, Ka-Ho and Gursoy, Emre and Truex, Stacey and Wu, Yanzhao},
  booktitle={2020 International Conference on Computing, Networking and Communications (ICNC)}, 
  title={Cross-Layer Strategic Ensemble Defense Against Adversarial Examples}, 
  year={2020},
  volume={},
  number={},
  pages={456-460},
  doi={10.1109/ICNC47757.2020.9049702}}

@article{dl-medical-image-analysis,
   author = "Shen, Dinggang and Wu, Guorong and Suk, Heung-Il",
   title = "Deep Learning in Medical Image Analysis", 
   journal= "Annual Review of Biomedical Engineering",
   year = "2017",
   volume = "19",
   number = "Volume 19",
   pages = "221-248",
   publisher = "Annual Reviews",
   issn = "1545-4274",
   type = "Journal Article",
  }

@INPROCEEDINGS{liu2019improving,
  author={Wu, Yanzhao and Liu, Ling},
  booktitle={2021 IEEE International Conference on Data Mining (ICDM)}, 
  title={Boosting Deep Ensemble Performance with Hierarchical Pruning}, 
  year={2021},
  volume={},
  number={},
  pages={1433-1438},
  doi={10.1109/ICDM51629.2021.00184}}

@article{golhani2018review,
title = {A review of neural networks in plant disease detection using hyperspectral data},
journal = {Information Processing in Agriculture},
volume = {5},
number = {3},
pages = {354-371},
year = {2018},
issn = {2214-3173},
author = {Kamlesh Golhani and Siva K. Balasundram and Ganesan Vadamalai and Biswajeet Pradhan}
}

@inproceedings{deep-ensembles,
 author = {Lakshminarayanan, Balaji and Pritzel, Alexander and Blundell, Charles},
 booktitle = {Advances in Neural Information Processing Systems},
 editor = {I. Guyon and U. Von Luxburg and S. Bengio and H. Wallach and R. Fergus and S. Vishwanathan and R. Garnett},
 pages = {},
 publisher = {Curran Associates, Inc.},
 title = {Simple and Scalable Predictive Uncertainty Estimation using Deep Ensembles},
 volume = {30},
 year = {2017}
}

@INPROCEEDINGS{Wu2023ModelLearning,
  author={Wu, Yanzhao and Chow, Ka-Ho and Wei, Wenqi and Liu, Ling},
  booktitle={2023 IEEE International Conference on Data Mining (ICDM)}, 
  title={Exploring Model Learning Heterogeneity for Boosting Ensemble Robustness}, 
  year={2023},
  volume={},
  number={},
  pages={648-657},
  doi={10.1109/ICDM58522.2023.00074}}

@ARTICLE{mohanty2016using,
AUTHOR={Mohanty, Sharada P.  and Hughes, David P.  and Salathé, Marcel },
TITLE={Using Deep Learning for Image-Based Plant Disease Detection},
JOURNAL={Frontiers in Plant Science},
VOLUME={7},
YEAR={2016},
ISSN={1664-462X},
}

@ARTICLE{fang2015current,
  author={Rayhana, Rakiba and Ma, Zhenyu and Liu, Zheng and Xiao, Gaozhi and Ruan, Yuefeng and Sangha, Jatinder S.},
  journal={IEEE Transactions on AgriFood Electronics}, 
  title={A Review on Plant Disease Detection Using Hyperspectral Imaging}, 
  year={2023},
  volume={1},
  number={2},
  pages={108-134},
  doi={10.1109/TAFE.2023.3329849}}

@article{too2019comparative,
title = {A comparative study of fine-tuning deep learning models for plant disease identification},
journal = {Computers and Electronics in Agriculture},
volume = {161},
pages = {272-279},
year = {2019},
note = {BigData and DSS in Agriculture},
issn = {0168-1699},
author = {Edna Chebet Too and Li Yujian and Sam Njuki and Liu Yingchun},
}

@INPROCEEDINGS{pang2019improving,
  title = 	 {Improving Adversarial Robustness via Promoting Ensemble Diversity},
  author =       {Pang, Tianyu and Xu, Kun and Du, Chao and Chen, Ning and Zhu, Jun},
  booktitle = 	 {Proceedings of the 36th International Conference on Machine Learning},
  pages = 	 {4970--4979},
  year = 	 {2019},
  editor = 	 {Chaudhuri, Kamalika and Salakhutdinov, Ruslan},
  volume = 	 {97},
  series = 	 {Proceedings of Machine Learning Research},
  month = 	 {09--15 Jun},
  publisher =    {PMLR}
}

@article{wu2024hierarchical,
author = {Wu, Yanzhao and Chow, Ka-Ho and Wei, Wenqi and Liu, Ling},
title = {Hierarchical Pruning of Deep Ensembles with Focal Diversity},
year = {2024},
issue_date = {February 2024},
publisher = {Association for Computing Machinery},
address = {New York, NY, USA},
volume = {15},
number = {1},
issn = {2157-6904},
url = {https://doi.org/10.1145/3633286},
doi = {10.1145/3633286},
journal = {ACM Trans. Intell. Syst. Technol.},
month = jan,
articleno = {15},
numpages = {24}
}

@INPROCEEDINGS{huang2017densely,
author = {Huang, Gao and Liu, Zhuang and van der Maaten, Laurens and Weinberger, Kilian Q.},
title = {Densely Connected Convolutional Networks},
booktitle = {Proceedings of the IEEE Conference on Computer Vision and Pattern Recognition (CVPR)},
month = {July},
year = {2017}
}

@INPROCEEDINGS{he2016deep,
author = {He, Kaiming and Zhang, Xiangyu and Ren, Shaoqing and Sun, Jian},
title = {Deep Residual Learning for Image Recognition},
booktitle = {Proceedings of the IEEE Conference on Computer Vision and Pattern Recognition (CVPR)},
month = {June},
year = {2016}
}

@article{Arsenovic2019,
AUTHOR = {Arsenovic, Marko and Karanovic, Mirjana and Sladojevic, Srdjan and Anderla, Andras and Stefanovic, Darko},
TITLE = {Solving Current Limitations of Deep Learning Based Approaches for Plant Disease Detection},
JOURNAL = {Symmetry},
VOLUME = {11},
YEAR = {2019},
NUMBER = {7},
ARTICLE-NUMBER = {939},
URL = {https://www.mdpi.com/2073-8994/11/7/939},
ISSN = {2073-8994},
DOI = {10.3390/sym11070939}
}

@article{ferentinos2018deep,
title = {Deep learning models for plant disease detection and diagnosis},
journal = {Computers and Electronics in Agriculture},
volume = {145},
pages = {311-318},
year = {2018},
issn = {0168-1699},
author = {Konstantinos P. Ferentinos}
}

@Article{dhaka2023role,
AUTHOR = {Dhaka, Vijaypal Singh and Kundu, Nidhi and Rani, Geeta and Zumpano, Ester and Vocaturo, Eugenio},
TITLE = {Role of Internet of Things and Deep Learning Techniques in Plant Disease Detection and Classification: A Focused Review},
JOURNAL = {Sensors},
VOLUME = {23},
YEAR = {2023},
NUMBER = {18},
ARTICLE-NUMBER = {7877},
PubMedID = {37765934},
ISSN = {1424-8220},
DOI = {10.3390/s23187877}
}

@Article{saleem2019plant,
AUTHOR = {Saleem, Muhammad Hammad and Potgieter, Johan and Arif, Khalid Mahmood},
TITLE = {Plant Disease Detection and Classification by Deep Learning},
JOURNAL = {Plants},
VOLUME = {8},
YEAR = {2019},
NUMBER = {11},
ARTICLE-NUMBER = {468},
PubMedID = {31683734},
ISSN = {2223-7747},
DOI = {10.3390/plants8110468}
}

@ARTICLE{li2021plant,
  author={Li, Lili and Zhang, Shujuan and Wang, Bin},
  journal={IEEE Access}, 
  title={Plant Disease Detection and Classification by Deep Learning—A Review}, 
  year={2021},
  volume={9},
  number={},
  pages={56683-56698},
  doi={10.1109/ACCESS.2021.3069646}}

@article{shoaib2023advanced,  
AUTHOR={Shoaib, Muhammad  and Shah, Babar  and EI-Sappagh, Shaker  and Ali, Akhtar  and Ullah, Asad  and Alenezi, Fayadh  and Gechev, Tsanko  and Hussain, Tariq  and Ali, Farman },
TITLE={An advanced deep learning models-based plant disease detection: A review of recent research},
JOURNAL={Frontiers in Plant Science},
VOLUME={Volume 14 - 2023},
YEAR={2023},
URL={https://www.frontiersin.org/journals/plant-science/articles/10.3389/fpls.2023.1158933},
DOI={10.3389/fpls.2023.1158933},
ISSN={1664-462X},
}

@incollection{debellis2021advances,
  title={Applications of deep learning in agriculture},
  author={Tripathi, Padmesh and Kumar, Nitendra and Rai, Mritunjay and Khan, Ayoub},
  booktitle={Artificial intelligence applications in agriculture and food quality improvement},
  pages={17--28},
  year={2022},
  publisher={IGI Global Scientific Publishing}
}

@article{kuncheva2003measures,
author = {Kuncheva, Ludmila and Whitaker, Chris},
year = {2003},
month = {05},
pages = {181-207},
title = {Measures of Diversity in Classifier Ensembles and Their Relationship with the Ensemble Accuracy},
volume = {51},
journal = {Machine Learning},
doi = {10.1023/A:1022859003006}
}

@inproceedings{skalak1996sources,
  title={The Sources of Increased Accuracy for Two Proposed Boosting Algorithms},
  author={David B. Skalak},
  booktitle={AAAI Conference on Artificial Intelligence},
  year={1996},
  url={https://api.semanticscholar.org/CorpusID:16135270}
}

@ARTICLE{Cohen1960,
  title={A Coefficient of Agreement for Nominal Scales},
  author={Cohen, Jacob},
  journal={Educational and Psychological Measurement},
  volume={20},
  number={1},
  pages={37--46},
  year={1960},
  publisher={SAGE Publications},
  doi={10.1177/001316446002000104}
}

@Article{Yule1900,
AUTHOR = {Eroglu, Deniz and Boghosian, Bruce M. and Borges, Ernesto P. and Tirnakli, Ugur},
TITLE = {The Statistics of q-Statistics},
JOURNAL = {Entropy},
VOLUME = {26},
YEAR = {2024},
NUMBER = {7},
ARTICLE-NUMBER = {554},
URL = {https://www.mdpi.com/1099-4300/26/7/554},
PubMedID = {39056916},
ISSN = {1099-4300},
DOI = {10.3390/e26070554}
}

@article{partridge1997software,
author = {Tang, E. and Suganthan, Ponnuthurai and Yao, Xin},
year = {2006},
month = {10},
pages = {247-271},
title = {{An Analysis of Diversity Measures}},
volume = {65},
journal = {Machine Learning},
doi = {10.1007/s10994-006-9449-2}
}

@INPROCEEDINGS{Bajait2020,
  author={Bajait, Vaishali and Malarvizhi, N.},
  booktitle={2020 4th International Conference on Electronics, Communication and Aerospace Technology (ICECA)}, 
  title={Review on Different Approaches for Crop Prediction and Disease Monitoring Techniques}, 
  year={2020},
  volume={},
  number={},
  pages={1244-1249},
  doi={10.1109/ICECA49313.2020.9297474}}

@article{kc-plant-pathology,
author = {Wang, Nian and Sundin, George W. and Fuente, Leonardo De La and Cubero, Jaime and Tatineni, Satyanarayana and Brewer, Marin T. and Zeng, Quan and Bock, Clive H. and Cunniffe, Nik J. and Wang, Congli and Candresse, Thierry and Chappell, Thomas and Coleman, Jeffrey J. and Munkvold, Gary},
title = {Key Challenges in Plant Pathology in the Next Decade},
journal = {Phytopathology},
volume = {114},
number = {5},
pages = {837-842},
year = {2024},
}

@article{review-detecting-plant-diseases,
title = {A review of advanced techniques for detecting plant diseases},
journal = {Computers and Electronics in Agriculture},
volume = {72},
number = {1},
pages = {1-13},
year = {2010},
issn = {0168-1699},
author = {Sindhuja Sankaran and Ashish Mishra and Reza Ehsani and Cristina Davis}
}

@article{ganaie2021ensemble,
author = {Ganaie, Mudasir and Hu, Minghui and Malik, Ashwani Kumar and Tanveer, M. and Suganthan, Ponnuthurai},
year = {2022},
month = {10},
pages = {105151},
title = {Ensemble deep learning: A review},
volume = {115},
journal = {Engineering Applications of Artificial Intelligence},
doi = {10.1016/j.engappai.2022.105151}
}

@INPROCEEDINGS{EnsembleBenchCogMI,
  author={Wu, Yanzhao and Liu, Ling and Xie, Zhongwei and Bae, Juhyun and Chow, Ka-Ho and Wei, Wenqi},
  booktitle={2020 IEEE Second International Conference on Cognitive Machine Intelligence (CogMI)}, 
  title={Promoting High Diversity Ensemble Learning with EnsembleBench}, 
  year={2020},
  volume={},
  number={},
  pages={208-217},
  doi={10.1109/CogMI50398.2020.00034}}

@INPROCEEDINGS{EnsembleBenchCVPR,
  author={Wu, Yanzhao and Liu, Ling and Xie, Zhongwei and Chow, Ka-Ho and Wei, Wenqi},
  booktitle={2021 IEEE/CVF Conference on Computer Vision and Pattern Recognition (CVPR)}, 
  title={Boosting Ensemble Accuracy by Revisiting Ensemble Diversity Metrics}, 
  year={2021},
  volume={},
  number={},
  pages={16464-16472},
  doi={10.1109/CVPR46437.2021.01620}}

@INPROCEEDINGS{sq-diversity-cogmi,
  author={Jin, Hongpeng and Aghdam, Maryam Akhavan and Nath Chowdary Medikonduru, Sai and Wei, Wenqi and Wang, Xuyu and Zhang, Wenbin and Wu, Yanzhao},
  booktitle={2024 IEEE 6th International Conference on Cognitive Machine Intelligence (CogMI)}, 
  title={Effective Diversity Optimizations for High Accuracy Deep Ensembles}, 
  year={2024},
  volume={},
  number={},
  pages={278-287},
  doi={10.1109/CogMI62246.2024.00044}}

@misc{dobrovsky2023dataset,
  author       = {Dobrovsky, Aline},
  title        = {{Plant Disease Classification - Merged Dataset}},
  year         = {2023},
  publisher    = {Kaggle},
  howpublished = {\url{https://www.kaggle.com/datasets/alinedobrovsky/plant-disease-classification-merged-dataset}},
  note         = {Accessed on 2025-01-19}
}

@misc{malviya2023dataset,
  author       = {Malviya, Deep},
  title        = {{New Plant Diseases Dataset (Augmented)}},
  year         = {2022},
  publisher    = {Kaggle},
  howpublished = {\url{https://www.kaggle.com/code/deepmalviya7/plant-disease-detection-using-cnn-with-96-84/input}},
  note         = {Accessed on 2025-01-19}
}

@INPROCEEDINGS{Efficientnet,
  title = 	 {{E}fficient{N}et: Rethinking Model Scaling for Convolutional Neural Networks},
  author =       {Tan, Mingxing and Le, Quoc},
  booktitle = 	 {Proceedings of the 36th International Conference on Machine Learning},
  pages = 	 {6105--6114},
  year = 	 {2019},
  editor = 	 {Chaudhuri, Kamalika and Salakhutdinov, Ruslan},
  volume = 	 {97},
  series = 	 {Proceedings of Machine Learning Research},
  month = 	 {09--15 Jun},
  publisher =    {PMLR},
  url = 	 {https://proceedings.mlr.press/v97/tan19a.html}
}

@inproceedings{VT2021,
title={An Image is Worth 16x16 Words: Transformers for Image Recognition at Scale},
author={Alexey Dosovitskiy and Lucas Beyer and Alexander Kolesnikov and Dirk Weissenborn and Xiaohua Zhai and Thomas Unterthiner and Mostafa Dehghani and Matthias Minderer and Georg Heigold and Sylvain Gelly and Jakob Uszkoreit and Neil Houlsby},
booktitle={International Conference on Learning Representations},
year={2021},
}

@INPROCEEDINGS{liu2019deep,
  author={Liu, Ling and Wei, Wenqi and Chow, Ka-Ho and Loper, Margaret and Gursoy, Emre and Truex, Stacey and Wu, Yanzhao},
  booktitle={2019 IEEE 16th International Conference on Mobile Ad Hoc and Sensor Systems (MASS)}, 
  title={Deep Neural Network Ensembles Against Deception: Ensemble Diversity, Accuracy and Robustness}, 
  year={2019},
  volume={},
  number={},
  pages={274-282},
  doi={10.1109/MASS.2019.00040}}

@article{banfield2005ensemble,
title = {Ensemble diversity measures and their application to thinning},
journal = {Information Fusion},
volume = {6},
number = {1},
pages = {49-62},
year = {2005},
note = {Diversity in Multiple Classifier Systems},
issn = {1566-2535},
author = {Robert E. Banfield and Lawrence O. Hall and Kevin W. Bowyer and W.Philip Kegelmeyer}
}
}
\end{document}